\documentclass[lettersize,journal]{IEEEtran}
\usepackage{amsmath,amsfonts}
\usepackage{algorithmic}
\usepackage{algorithm}
\usepackage{array}
\usepackage[caption=false,font=normalsize,labelfont=sf,textfont=sf]{subfig}
\usepackage{textcomp}
\usepackage{stfloats}
\usepackage{url}
\usepackage{verbatim}
\usepackage{graphicx}
\usepackage{cite}
\usepackage{multirow}
\usepackage{bm}
\usepackage{amssymb}
\usepackage{amsmath}

\usepackage{color}

\begin{document}

\title{Single-Stage Broad Multi-Instance Multi-Label Learning (BMIML) with Diverse Inter-Correlations and its Application to Medical Image Classification}

\author{Qi Lai,
        Jianhang Zhou,
        Yanfen Gan,
        Chi-Man Vong,~\IEEEmembership{Senior Member,~IEEE,}
        
        and C.L. Philip Chen,~\IEEEmembership{Fellow,~IEEE}
        
\thanks{© 2023 IEEE.  Personal use of this material is permitted.  Permission from IEEE must be obtained for all other uses, in any current or future media, including reprinting/republishing this material for advertising or promotional purposes, creating new collective works, for resale or redistribution to servers or lists, or reuse of any copyrighted component of this work in other works.}}


\maketitle

\begin{abstract}

In many real-world applications, one object (e.g., image) can be represented or described by multiple instances (e.g., image patches) and simultaneously associated with multiple labels. Such applications can be formulated as \textit{multi-instance multi-label learning} (MIML) problems and have been extensively studied during the past few years. Existing MIML methods are useful in many applications but most of which suffer from relatively low accuracy and training efficiency due to several issues: i) \textit{the inter-label correlations} (i.e., the probabilistic correlations between the multiple labels corresponding to an object) are neglected; ii) \textit{the inter-instance correlations} (i.e., the probabilistic correlations of different instances in predicting the object label) cannot be learned \textit{directly} (or jointly) with other types of correlations due to the missing instance labels; iii) diverse inter-correlations (e.g., inter-label correlations, inter-instance correlations) can only be learned in multiple stages. To resolve these issues, a new single-stage framework called \textit{broad multi-instance multi-label learning} (BMIML) is proposed. In BMIML, there are three innovative modules: i) an \textit{auto-weighted label enhancement learning} (AWLEL) based on broad learning system (BLS) is designed, which simultaneously and efficiently captures the inter-label correlations while traditional BLS cannot; ii) A specific MIML neural network called \textit{scalable multi-instance probabilistic regression} (SMIPR) is constructed to effectively estimate the inter-instance correlations using the object label only, which can provide additional probabilistic information for learning; iii) Finally, an \textit{interactive decision optimization} (IDO) is designed to combine and optimize the results from AWLEL and SMIPR and form a single-stage framework. As a result, BMIML can achieve simultaneous learning of diverse inter-correlations between whole images, instances, and labels in single stage for higher classification accuracy and much faster training time. In this work, medical image classifications is employed as an illustration. Experiments show that BMIML is highly competitive to (or even better than) existing methods in accuracy and much faster than most MIML methods even for large medical image data sets ($>$ 90K images).
\end{abstract}


\section{Introduction}
\IEEEPARstart{I}{n} traditional supervised learning, one object is only represented by a single instance and associated with a single label. However, in many real-world applications, one object can be naturally described by a collection of instances (called a \textit{bag}) and has multiple class labels simultaneously. Such applications can be formulated as \textit{multi-instance mult-label learning} (MIML) problem \cite{zhou2006multi} and have been extensively applied in many fields such as image classification \cite{jie2017image,song2018deep}, video annotation \cite{zhang2020multi,biswas2021multiple}, biomedicine \cite{li2018hmiml} and protein function prediction \cite{wu2014genome}. Out of many MIML applications, medical image classification is one of the most popular research areas for its practical use. Nowadays, with higher pressure on public health and a shortage of professionals on different types of medical imaging \cite{chang2020synthetic}, it is necessary to further investigate general, effective, and efficient automated methods for clinical use. In recent works \cite{zhao20213d,zhou2019high,li2021volumenet}, extensive applications have been proposed to explore a possible way of automated disease classification. Literature \cite{xu2020multi,li2021dual,ji2021learning} show that the MIML-based methods have great potential in automated disease classification and clinical diagnosis in the medical field. These indicate the feasibility of a MIML-based automated approach for disease classification. Therefore, medical image classification is employed as an illustrative application in this work.

\begin{figure}[!t]
\centering
\includegraphics[width=3.5in]{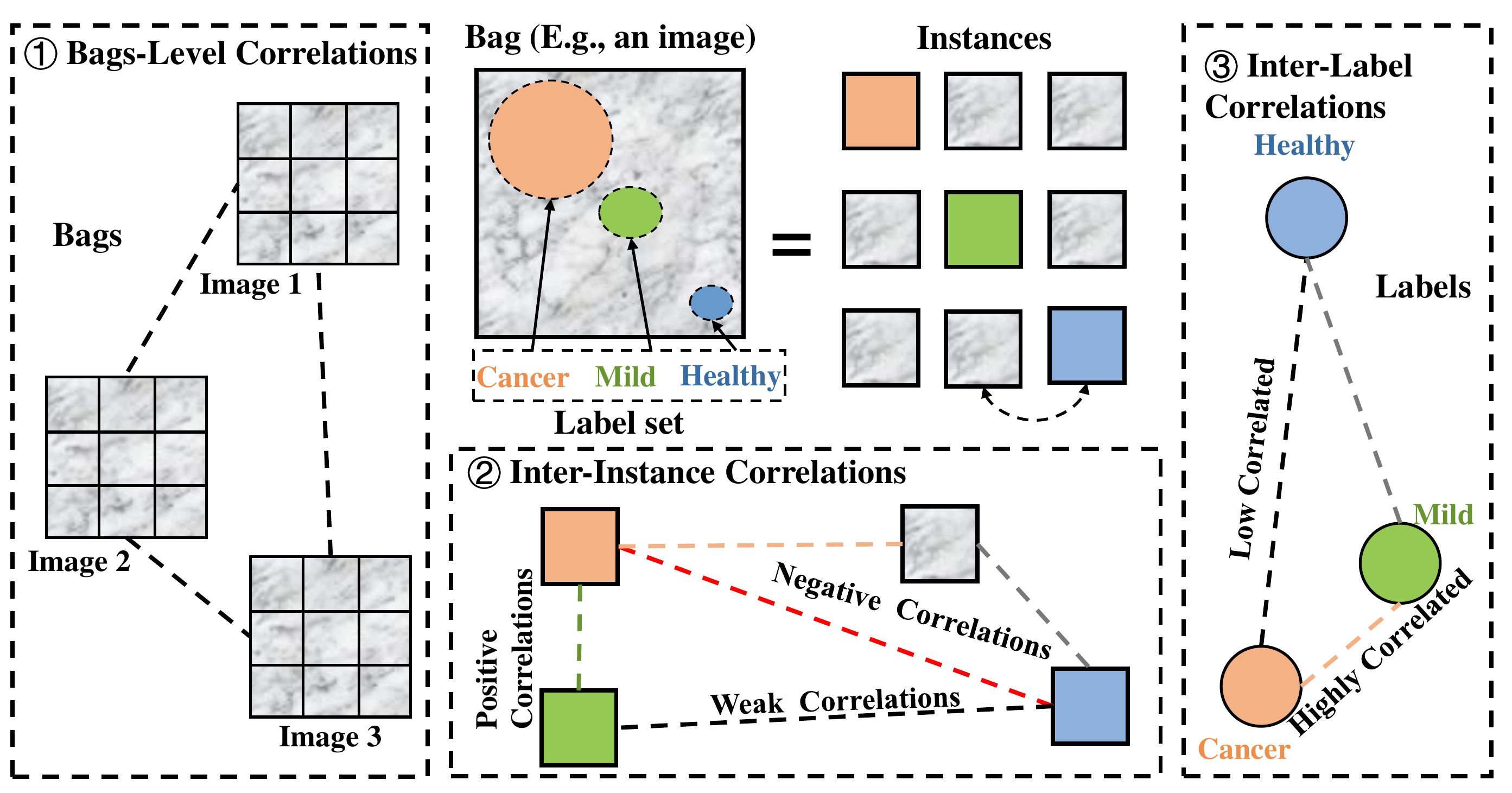}
\caption{The details of diverse correlations between bags (\textit{global view}), instances (\textit{local view}), and the multiple labels. Solid lines indicate the correlations between bags, instances, and labels. Dash lines indicate the inter-correlations (partially ignored in existing works but practically all are required) between bags-bags, instances-instances, labels-labels.}
\label{fig:diverse_correlation}
\end{figure}

Medical image classification task is typically formulated as a multi-class or \textit{multi-label learning} (MLL) problem. Strictly speaking, the medical image is usually multi-labeled, and for each image, the distribution of different labels is often imbalanced. As shown in Fig. \ref{fig:diverse_correlation}, \textit{Label} 1 is the dominant position and is accurately predicted while \textit{Label 3} is almost ignored since \textit{Label 3} only occupies a small part of the images. For this reason, the easily recognized labels usually result in a dominant position, which always leads to relatively poor performance. MIML has been applied to deal with the above problem, which offers a way for understanding the correlations between the input images and the output labels. In MIML setting, an image can be divided into several segments or patches (i.e., \textit{instances}) so that the multi-label classification tasks can be performed at the instance-level as shown in Fig. \ref{fig:diverse_correlation}. Meanwhile, a collection of instances is called a \textit{bag} which can represent an image (training sample) and the bag is assigned with a set of multiple class labels (i.e., \textit{label set}).

Practically, clinicians consider diverse correlations in medical image classification as illustrated in Fig. \ref{fig:diverse_correlation}, where the solid lines indicate the correlations between bags (\textit{global view}), instances (\textit{local view}), and labels while the dash lines indicate the inter-correlations (partially ignored in existing works but practically all are required) between \textit{bags-bags}, \textit{instances-instances}, \textit{labels-labels}. In other words, all correlations are practically used to estimate the correlation between bags and labels to achieve the best possible classification performance \cite{wang2021semi}. For example, to identify if an image is relevant to suspicion lesions (i.e., the correlation between bags and labels) or not, the following information should be considered simultaneously: 

\begin{enumerate}
    \item the bag-level correlations that reveal the difference of medical images of distinct diseases from the perspective of the whole image (global view);
    \item the inter-instance correlations that reveal which parts of an image are significant to distinguish from different diseases (local view);
    \item the inter-label correlations that quantitatively indicate the margin between two diseases with the class probabilities (or confidence) of the intraclass samples.
\end{enumerate}

Therefore, for more effective medical image classification, the diverse correlations which were partially \textit{neglected} in existing methods should be simultaneously considered \cite{xing2019multi}.

However, how to make use of inter-instance correlations \cite{li2015boosting, chi2021explicit} (i.e., the probabilistic correlations of different instances in predicting the bag labels) remains a challenging research topic because, in almost all available data sets, only image/bag-level (global view) labels are available while instance-level (local view) labels are missing due to the heavy burden in manual labeling for the clinicians. For this reason, traditional supervised learning methods are \textit{unable} to learn the inter-instance correlations \textit{directly}.

Although existing MIML methods can learn the inter-instance correlations \textit{indirectly} through multiple independent learning procedures, this indirect multi-stage way will affect the model performance in accuracy and efficiency. Moreover, considering diverse correlations will bring time-consuming which is another challenge for existing MIML methods \cite{huang2018fast}, especially in large data sets. Thus, it is necessary to design a unified single-stage interactive framework that can learn the information of whole images/bags (global view) and instances (local view) simultaneously and improve efficiency. However, existing MIML methods do not provide the way of simultaneous learning so that it becomes highly nontrivial to implement over MIML methods. To our best knowledge, there is no such simultaneous learning mechanism of diverse correlations for MIML in existing works as summarized in Table \ref{tab:table1}.

Recently, efficient discriminative learning called \textit{Broad Learning System} (BLS) \cite{chen2017broad} was proposed. The main advantage of BLS is its efficient network training under random feature mapping with the ability to jointly learning of multiple sub-networks. In BLS, the original inputs are transferred as the mapping features and placed in the \textit{feature layer} (a sub-network), and the structure is extended to the \textit{enhancement layer} (another sub-network) in a broad sense. Both the feature and enhancement layers are then connected to the output layers. Thus, BLS offers the necessary mechanism of simultaneous/joint learning efficiently.

Although BLS can deal with MLL tasks (e.g., one sample corresponding to several labels), it does not consider the inter-label correlations \cite{nguyen2021modular, ma2020latent} which must be considered in MLL. Moreover, BLS requires that all inputs are independent of each other and simply sets the entire data matrix X as input \cite{chen2017broad}. However, medical image classification is always a MIML problem in which the instances of the input samples are highly relevant, so it is impossible to assume all inputs independently. Therefore, the application of BLS in medical image classification becomes nontrivial and challenging.

\begin{table*}[!t]
\caption{Diverse inter-correlations exploited by proposed BMIML and MIML methods\label{tab:table1}}
\centering
\begin{tabular}{lccccccc}
\hline
\multicolumn{1}{c}{\multirow{2}{*}{Approaches}} & \multicolumn{6}{c}{Diverse Inter-correlations}\\ \cline{2-7} 
\multicolumn{1}{c}{}                            & \multicolumn{1}{c}{Bag-Bag} & \multicolumn{1}{c}{Inter-instances} & \multicolumn{1}{c}{Inter-labels} & \multicolumn{1}{c}{Bag-Instance} & \multicolumn{1}{c}{Bag-Label} & \multicolumn{1}{c}{Instance-Label} \\ \hline
MIMLNN{\cite{zhou2012multi}}                                  &                             & $\surd$                             &                                  & $\surd$                          & $\surd$                       &                                    \\ \hline
MIMLSVM{\cite{zhou2006multi}}                                 & $\surd$                     &                                     &                                  & $\surd$                          & $\surd$                       &                                    \\ \hline
MIMLmiSVM{\cite{zhou2012multi}}                                & $\surd$                     &                                     &                                  & $\surd$                          & $\surd$                       &                                    \\ \hline
MIMLkNN{\cite{zhang2010k}}                                 & $\surd$                     &                                     &                                  & $\surd$                          & $\surd$                       & $\surd$                            \\ \hline
MIMLBoost{\cite{zhou2006multi}}                               & $\surd$                     &                                     &                                  & $\surd$                          & $\surd$                       &                                    \\ \hline
MIMLfast{\cite{huang2018fast}}                                &                             &                                     & $\surd$                          & $\surd$                          & $\surd$                       & $\surd$                            \\ \hline
DeepMIML{\cite{feng2017deep}}                                & $\surd$                     & $\surd$                             &                                  & $\surd$                          & $\surd$                       & $\surd$                            \\ \hline
\textbf{Proposed BMIML}                         & $\surd$                     & $\surd$                             & $\surd$                          & $\surd$                          & $\surd$                       & $\surd$                            \\ \hline
\end{tabular}
\end{table*}

In this paper, a novel approach for medical image classification called Broad Multi-Instance Multi-Label network (BMIML) is proposed. Concretely, the BMIML is based on BLS which can jointly learn multiple sub-networks in a broad sense so that the diverse correlations between bags, instances, and labels can be simultaneously captured. However, standard BLS cannot capture the inter-label correlation which is necessary for handling MIML problems. Also, it cannot utilize the inter-instance correlations for training \textit{directly}. Thus, in BMIML, an interactive framework is newly designed that includes three novel modules: i) \textit{auto-weighted label enhancement learning} (AWLEL), ii) \textit{scalable multi-instance probabilistic regression} (SMIPR), and iii) an \textit{interactive decision optimization} (IDO). On the one hand, AWLEL as part of MLL in BMIML can model diverse correlations, including the inter-label correlation which can improve the accuracy of BMIML. On the other hand, SMIPR as part of multi-instance learning in BMIML is a way to model the inter-instance correlations using bag labels only. Finally, IDO works as a bridge to connect AWLEL and SMIPR to integrate their results into a network, forming an interactive single-stage framework that can deal with MIML problems efficiently and effectively. 

Hence BMIML overcomes the weaknesses of the BLS and existing MIML framework by simultaneously learning the diverse (inter-)correlations. The illustrations of the diverse (inter-)correlations mentioned above are shown in Fig. \ref{fig:diverse_correlation} and Table \ref{tab:table1}. The main contributions of BMIML are summarized below:

\begin{enumerate}
    \item Through our proposed method BMIML, the diverse correlations between bags, instances, and multiple labels can be considered simultaneously for higher classification accuracy. This simultaneous consideration of diverse correlations cannot be done in existing MIML works as illustrated in Table \ref{tab:table1}.
    \item In BMIML, an interactive single-stage learning framework is newly designed which can simultaneously consider the correlations of both global views (whole images/bags) and local view (image patches/instances) for image classification tasks. This single-stage framework can further improve classification accuracy, learning efficiency, and human burden, especially on large data sets. This is a non-trivial challenging task because local view labels are always missing in the training data set.
\end{enumerate}

The organization of this article is as below. Section \ref{sec:pre}  provides a brief review of BLS and MIML. Section \ref{sec:bmiml} details our proposed methods: BMIML, including AWLEL, SMIPR, and IDO. Section \ref{sec:exp} demonstrates the experimental results with analysis and discussion. At last, a conclusion is drawn in Section \ref{sec:conclusion}.

\section{Preliminaries} \label{sec:pre}

\subsection{Multi-instance Multi-label Learning (MIML)} \label{ssec:miml}

\begin{figure}[!ht]
\centering
\includegraphics[width=3.5in]{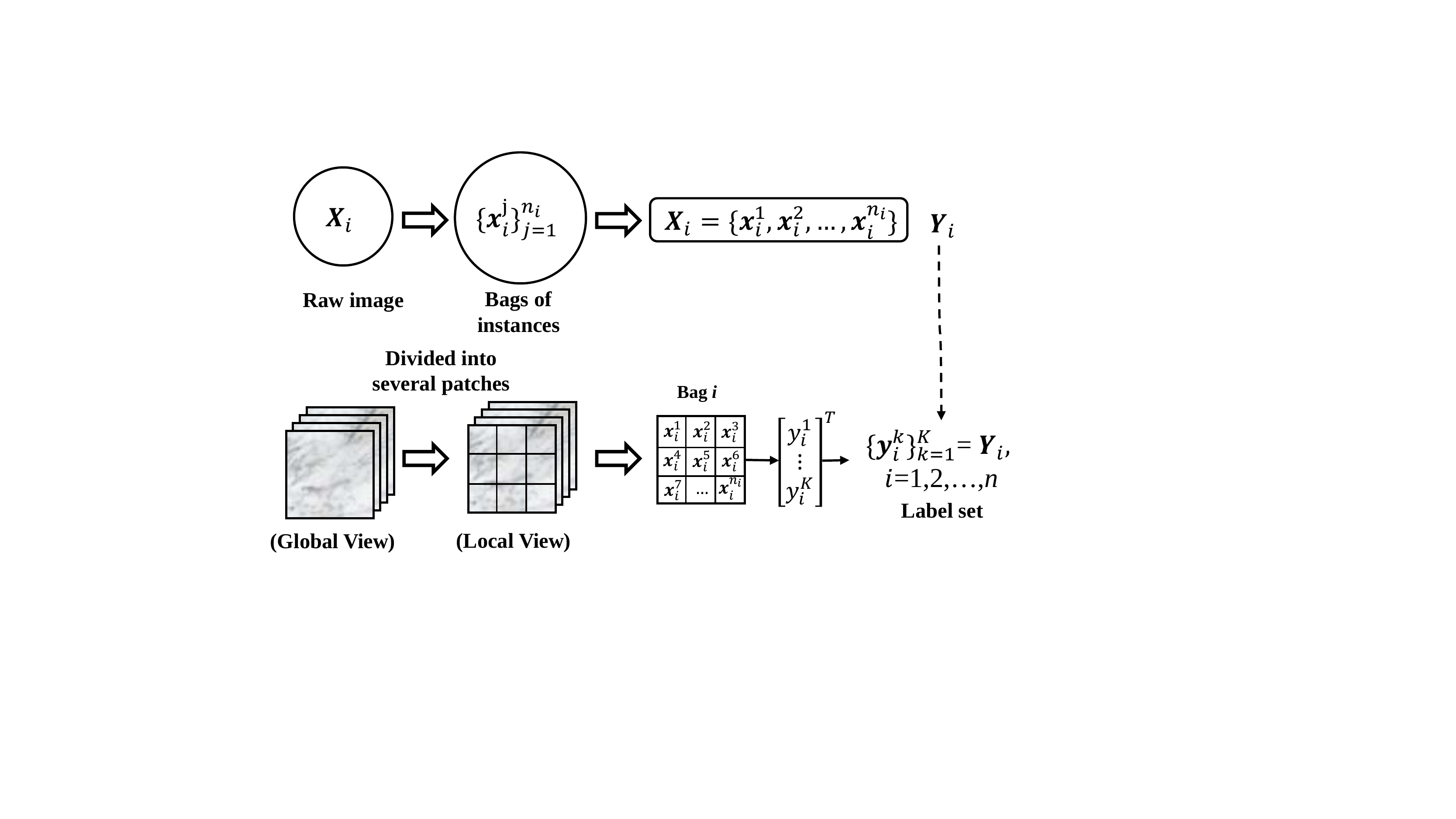}
\caption{A brief introduction of the MIML setting. $\mathbf{X}_i$ represents the $i$-th image in the dataset and then it was divided into several patches called instances $\mathbf{x}_i^j$, where $j=1,2,…,n_i$, $n_i$ is the number of the instances in the $i$-th image. $\mathbf{Y}_i $is the label set associated with $\mathbf{X}_i$. $K$ indicates the number of categories.}
\label{fig:introduce_miml}
\end{figure}

Clinically, a medical image can be described by multiple semantic labels, as shown in Fig. \ref{fig:diverse_correlation}. However, these labels are only closely related to their respective \textit{regions/patches} (called \textit{instances}) rather than the entire image \cite{li2021multi}, as illustrated in Fig. \ref{fig:diverse_correlation}. For this reason, a more rational and natural strategy is to model medical image classification as a \textit{multi-instance multi-label learning }(MIML) tasks \cite{zhou2012multi}. As illustrated in Fig.\ref{fig:introduce_miml}, given a training set $\left \{ (\bm{X}_i,\bm{Y}_i ) \right \}_{i=1}^n$ where $\bm{X}_i= \left \{ \bm{x}_i^1,\bm{x}_i^2,\dots,\bm{x}_i^{n_i} \right \} (i=1 ,2,\dots,n)$ represents a bag of instances (image patches) $\bm{x}_i^j$ divided from the $i^{th}$ original image $\bm{X}_i$, $\bm{Y}_i$ is a $K$-dimensional label vector $[y_i^1,y_i^2,\dots,y_i^K]$ or $\bm{X}_i$ and $y_i^k \in \left \{ -1,1 \right \} ,k=1,2,\dots,K$ entry $y_i^k$ indicates the membership corresponding to $\bm{X}_i$ with the kth class label. Unfortunately, as shown in Fig.\ref{fig:introduce_miml}, the relation between $\bm{Y}_i$ and each instance $\bm{x}_i^j$ is not explicitly indicated in the training set, which is exactly our training target. Therefore, we introduce a probabilistic regression framework to construct the probabilistic correlations between instances $\bm{x}_i^j$ and bag label $\bm{Y}_i$. Based on the training set, the MIML probabilistic regression aims to approximate a function that can predict the class probability (or confidence) of testing set as accurately as possible.

\subsection{Broad Learning System (BLS)} \label{ssec:bls}

BLS is simply introduced here and the readers can refer \cite{chen2017broad} for details. Given the training set $\left \{ \bm{X},\bm{Y} \right \} \in \bm{R}^{N \times (D+K)}$ where $\bm{X}=[\bm{X}_i] \in \bm{R}^{N \times D}$ is the input matrix where $X_i$ denotes the $i^{th}$ sample with the relevant output $\bm{Y}_i$ and $\bm{Y} = [\bm{Y}_i] \in \bm{R}^{N \times K}$ is the output matrix. $D$ is the dimension of input vector $\bm{X}_i$ and $K$ is the number of class labels. Then the input matrix $\bm{X}$ is mapped into a series of random features  $\bm{Z}_{m_1}, m_1=1 \  to \  M_1$. Each feature mapping node $Z_{m_1}$ can be represented as:

\begin{equation} \label{eq:bls_z}
\bm{Z}_{m_1}=\xi_{m_1}^z (\bm{X}\bm{w}_{m_1}^z+\bm{\beta}_{m_1}^z)
\end{equation}
where $m_1$ is a user-specified parameter and $\xi_{m_1}^z$ is an activation function (e.g., sigmoid). $bm{w}_{m_1}^z$ and $\bm{\beta}_{m_1}^z$ are the randomly generated weights and bias matrices with the proper dimensions for input $\bm{X}$, respectively. Similarly, the enhancement nodes $\bm{H}_{m_2}, m_2=1 \  to \  M_2$ are denoted by:

\begin{equation} \label{eq:bls_h}
\bm{H}_{m_2}=\xi_{m_2}^h (\bm{Z}_{m_1}\bm{w}_{m_2}^h+\bm{\beta}_{m_2}^h )
\end{equation}
where $\xi_{m_2}^h$ is a non-linear function (e.g., $tanh(\cdot)$) which can be selected differently in building a model as well as $\xi_{m_1}^z$ and $m_2$ is a user-specified parameter. Here, the number of mapping nodes $\bm{Z}_{m_1}$ and enhancement nodes $\bm{H}_{m_2}$ can be same or different. It is set according to the actual situation and will not be described here. $\bm{w}_{m_2}^h$ and $\bm{\beta}_{m_2}^h$ are respectively random weights and bias matrices for the mapped features $\bm{Z}_{m_1}$. Hence, the output nodes $\bm{Y}$ of BLS can be denoted by:

\begin{equation} \label{eq:bls_y}
\bm{Y}=[\bm{Z}_1,\bm{Z}_2,\dots,\bm{Z}_{M_1},\bm{H}_1,\bm{H}_2,\dots,\bm{H}_{M_2} ]\bm{W}
\end{equation}
where the weights $W$ are connecting the layer of features nodes and the layer of enhancement nodes to the output nodes, and $W=A^{+}Y$, which can be easily computed using ridge regression approximation of pseudoinverse as follows:

\begin{equation} \label{eq:bls_op}
arg\min_{\bm{W}}\left \|\bm{A}\bm{W}-\bm{Y}\right\|_2^2+\lambda\left\|\bm{W} \right\|_2^2
\end{equation}

\begin{equation} \label{eq:bls_sol}
\bm{A}^+=\lim_{\lambda \to 0} \left ( \lambda \bm{I} + \bm{A}\bm{A}^T \right )^{-1}\bm{A}^T
\end{equation}
where $\bm{A}=[\bm{Z}_1,\bm{Z}_2,\dots,\bm{Z}_{M_1},\bm{H}_1,\bm{H}_2,\dots,\bm{H}_{M_2}]$. The value $\lambda$ indicates the further constraints on the squared weights $\bm{W}$. Consequently, we have

\begin{equation} \label{eq:bls_W}
\bm{W}=(\lambda\bm{I}+\bm{A}\bm{A}^T)^{-1} \bm{A}^T \bm{Y}
\end{equation}

Since BLS simply takes the entire data matrix $\bm{X}$ as input \cite{chen2017broad} i.e., all inputs are assumed independent of each other, and it cannot capture correlations between multiple labels. Therefore, BLS cannot directly employ MIML tasks. In our tasks, an improved BLS framework is designed by i) adding a retargeting layer enables BLS to capture the inter-label correlations; and ii) simultaneously modeling the diverse correlations between the bags, instances, and labels (see Table \ref{tab:table1}).

\begin{figure*}[!t]
\centering
\includegraphics[width=6.5in]{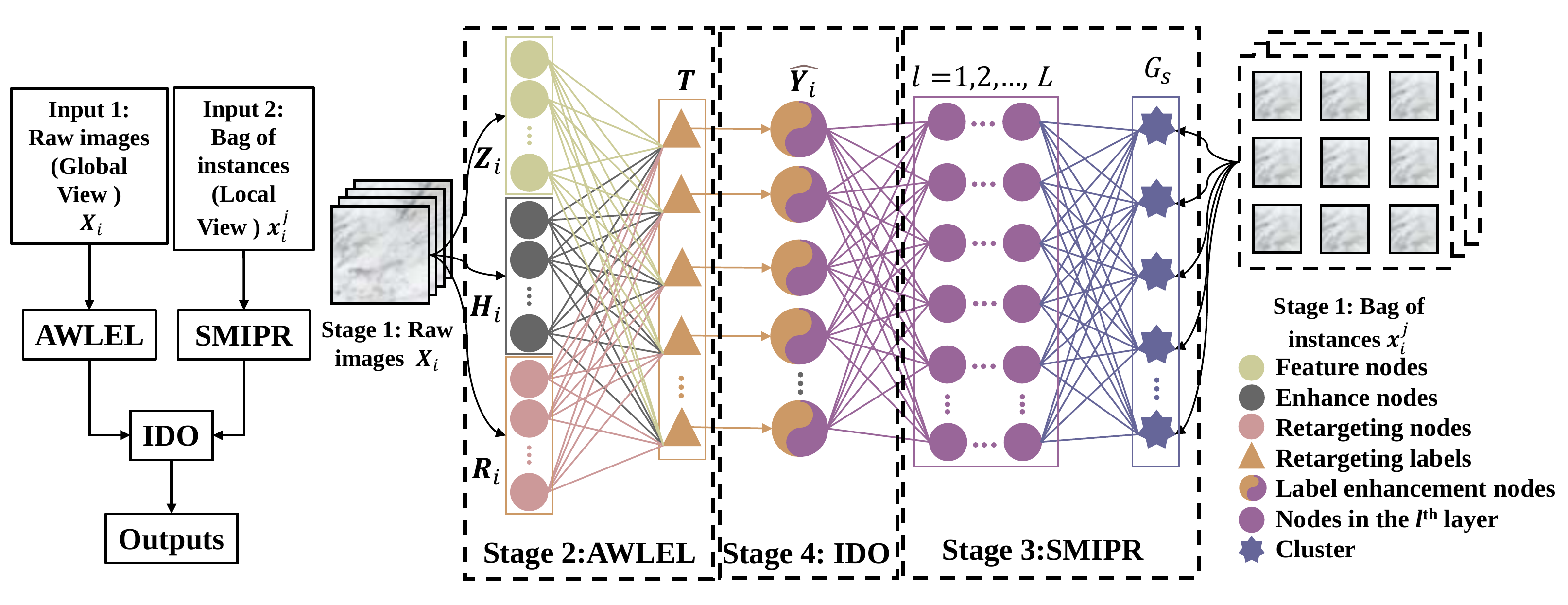}
\caption{Structure of the proposed BMIML.}
\label{fig:BMIML}
\end{figure*}

\section{Proposed BMIML} \label{sec:bmiml}
BLS is good at joint learning of different information and therefore suitable for learning diverse correlations simultaneously. Although BLS has demonstrated its strong classification ability in many fields \cite{chu2019weighted,zhao2020semi,ye2020adaptive}, it does not work very well for semantically complex images (e.g., multi-label images) sinceit cannot consider the inter-label correlations and the property of weakly discriminative features in the image. In this section, aiming at improving the performance for medical image classification, an single-stage interactive framework is newly designed based on i) \textit{auto-weighted label enhancement learning} (AWLEL) to process MLL in MIML, i.e., handling diverse correlations, and reformulating the original single-label space into an enhanced retargeted multi-label space by considering intraclass and interclass scatters for better discrimination under weak features (as shown in Fig. \ref{fig:target_gap}); ii) a novel \textit{scalable multi-instance probabilistic regression} (SMIPR) to provide multi-instance probabilistic predictions by fully utilizing the inter-instance correlations (as shown in Fig. 5); and iii) using an \textit{interactive decision optimization} (IDO) to combine the AWLEL and SMIPR, forming an end-to-end framework to deal with MIML tasks. The entire process of the proposed BMIML is summarized in Fig. \ref{fig:BMIML}, which has the following four computational stages.

\subsection{Overview of BMIML Stages} \label{ssec:overview}
\textit{Stage 1 (Preprocessing)}: The training data set includes original images $\bm{X}_i$ ({global view}) and instances $\bm{x}_i^j$ (local view, simply dividing from original images, detailed in Section \ref{ssec:settings}), which are the inputs to the AWLEL and SMIPR, respectively. 

\textit{Stage 2 (Auto-Weighted Label Enhancement Learning)}: AWLEL is designed based on the BLS, as shown in Fig.\ref{fig:BMIML}, stage 2. Different from the standard BLS, a new \textit{retargeting layer} $\bm{R}_i$ is added in the BLS that aims to improve the issue of MLL tasks and can automatically generate the retargeted labels for instances from bag-level labels. In addition, it can guarantee to impose the constraint of large margin of classification boundary for the requirement of correct classification for each data point. The learning details for the proposed AWLEL module is described in Section \ref{ssec:awlel}, and its optimization strategy is detailed in Section \ref{ssec:os}.

\textit{Stage 3 (Scalable Multi-Instance Probabilistic Regression)}: SMIPR is a neural network specifically designed for MIML which performs probabilistic regression on each instance according to the retargeted labels $\textbf{T}_i$ only generated by AWLEL. In other words, SMIPR can estimate the inter-instance correlations by using the images/bags labels\textit{ only}. Different from traditional neural network structure, the first layer of SMIPR is a clustering process to generate \textit{S disjoint groups of bags} ${G_1,G_2,\dots,G_S}$, and calculate the corresponding medoids $v_p$ of the clusters $G_p$, \textit{p = 1} to \textit{S}. Since clustering helps uncover the underlying structure of the training data set, the medoid of each cluster may make full use of the instance information and encodes some distribution information of different bags. The detail is discussed Section \ref{ssec:SMIPR}.

\textit{Stage 4 (Interactive Decision Optimization)}: In almost all data sets, there are only bag-level (\textit{global view}) labels while the instance-level (\textit{local view}) labels are missing. Therefore, an \textit{interactive decision optimization} (IDO) is designed as a bridge to connect the above two modules: AWLEL and SMIPR. In other words, IDO integrates the results of i) simultaneous learning of diverse correlations and ii) direct learning the instance class membership probability in a single network, which can achieve an end-to-end learning. The detailed is provided by Section \ref{ssec:ido}.

In summary, we aim to construct BMIML for the effective and efficient classification of medical images. For this purpose, an interactive end-to-end learning framework is designed, as shown in Figure 3. First, the AWLEL captures the inter-label correlation, which helps to enlarge the target gaps between the interclass samples. Then, SMIPR was employed to learn the inter-instance correlation according to the inter-label correlation so that it can better capture the local view information. Finally, the AWLEL and SMIPR are connected under IDO and therefore a single-stage muti-instance multi-label learning framework can be achieved. Also, for this reason, IDO cannot work independently for the multi-label image classification task.

\subsection{Auto-Weighted Label Enhancement Learning (AWLEL)} \label{ssec:awlel}
\begin{figure}[!h]
\centering
\includegraphics[width=3in]{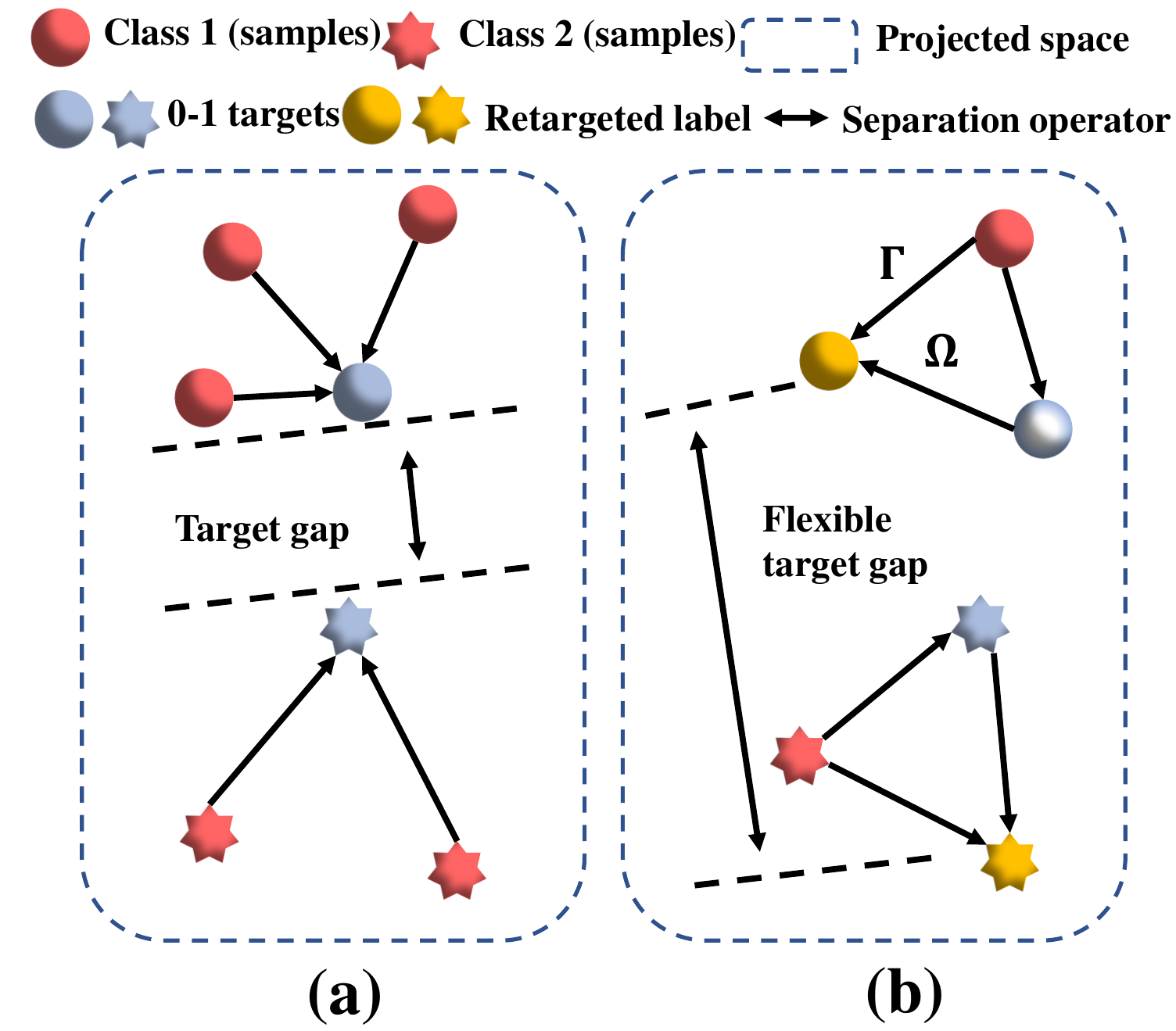}
\caption{The difference between the standard BLS and the AWLEL. In (a), all intraclass samples shrink to the fixed 0-1 targets in standard BLS projector space while in (b), AWLEL can auto-weight all intraclass samples to enlarge the gaps between interclass samples.}
\label{fig:target_gap}
\end{figure}

In standard BLS, all inputs are assumed independent of each other and the entire data matrix $\bm{X}$ is taken as input. Besides, the output matrix $\bm{Y}$ in standard BLS is a strict zero-one matrix, i.e., only the label entry of each row is one, where $label \in \left \{ 1,2,\dots,K \right \}$ is class label of sample $\bm{X_i}$, as shown in Fig. \ref{fig:target_gap}(a). Practically, a medical image is always associated with multiple labels and the distribution of labels is imbalanced. Strict zero-one indicator do not make sense and may be detrimental to classification. Moreover, a series of instances (divided from an image) in a bag are often dependent with each other (i.e., inter-instance correlations). Hence, there is also a probabilistic correlation between multiple labels (i.e., inter-label correlations) associated with a bag. To tackle this issue, another sub-network (called \textit{retargeting nodes}) is added into the standard BLS which can enable BLS for multi-label tasks and capture the inter-label correlations. In our work, the retargeting nodes is defined as

\begin{equation} \label{eq:ri}
\bm{R}_i=\xi_i^r \left (\bm{X}_i \bm{w}_{m_1}^z+\bm{Z}_{m_1} \bm{w}_{m_2}^h+\bm{\beta}_i^r \right )
\end{equation}
where $\xi_i^r$ is a non-linear function (e.g., \textit{Tribas}) and $\bm{\beta}_i^r$ is a regularization parameter controlling the degree of bias. The weights $\bm{w}_{m_1}^z$ and $\bm{w}_{m_2}^h$ can be generated from Eqs. (\ref{eq:bls_z}) and (\ref{eq:bls_h}), respectively. And then we define the \textit{retargeted labels} of the \textit{i}th training sample as below:

\begin{equation} \label{eq:ti}
\bm{T}_i=(\bm{X}_i,\bm{Z}_{m_1},\bm{H}_{m_2},\bm{R}_i)\bm{w}_i^t
\end{equation}
where the weight $\bm{w}_i^t$ is jointly optimized by feature nodes, enhance nodes, and retargeting nodes. The simultaneously learning of random mapping and regression target of BLS is as follows:

\begin{equation} \label{eq:wt}
arg\min_{\bm{w}^t,\bm{T}} \left \|\bm{Aw}^t-\bm{T} \right \|_2^2+\lambda\left \|\bm{w}^t\right\|_2^2
\end{equation}
where $\bm{T}\in\mathbb{R}^{N\times K}$ is the \textit{retargeted labels} and consists of $\bm{T}_i$ which can reflect the classification separability (see Fig.\ref{fig:BMIML} Stage 2) of each sample (global view) with respect to different class labels. To improve the interclass separability, Eq. (\ref{eq:wt}) is reformulated as:

\begin{equation} \label{eq:rwt}
\begin{aligned}
arg \min_{\bm{w}^t_i,\bm{T}_i} \sum_{i=1}^{N} & (  \gamma_i\left \| \bm{A}_i\bm{w}_i^t-\bm{T}_i \right \|_2^2+\lambda\left \| \bm{w}_i^t \right \|^2_2 \\
& + \theta \omega_i \left \| \bm{T}_i-\bm{Y}_i \right \|_2^2) 
\end{aligned}
\end{equation}
where the weighted penalty factors $\gamma_i$ and $\omega_i$ control the effect of outliers and the balance between the loss components in the total loss, $\gamma_i=\left ( \frac{1}{[\left \|\bm{A}_i\bm{W}_i-\bm{T}_i\right \| _2]} \right )$ and $omega_i=\left ( \frac{1}{[\left \|\bm{T}_i-\bm{Y}_i \right \| _2]} \right )$. $\bm{A}_i$ is the \textit{i}th row vector in the matrix $\bm{A}$, as illustrated in Section \ref{ssec:bls}. The value $\vartheta$ indicates further constraint on the squared difference of retargeted label and ground truth. Using the diagonal matrices $\bm{\Gamma} =\left [\gamma_1,\gamma_2,...,\gamma_N  \right ]^T$ and $\bm{\Omega} =\left [\omega_1,\omega_2,...,\omega_N \right ]^T$ and combining with Eq. (\ref{eq:rwt}), we have:

\begin{equation} \label{eq:wt2}
\begin{aligned}
arg\min_{\bm{w}^t,\bm{T}} & \left \|\sqrt{\bm{\Gamma}} \bm{Aw}^t -\bm{T} \right \|_2^2+\lambda\left \|\bm{w}^t\right\|_2^2\\
&+\vartheta \left \| \sqrt{\bm{\Omega}} \left ( \bm{T}-\bm{Y} \right )  \right \|_2^2
\end{aligned}
\end{equation}

As shown in Fig.\ref{fig:target_gap} (a) and (b), we aim to overcome the limitation of BLS and promote effective separability. Therefore, we expect the samples are drawn from the same class and gather to the corresponding targets. This allows adaptive learning of intraclass targets while enlarging the gaps between interclass targets, resulting to more generalized properties. As formulated in Eq. (\ref{eq:rwt}), unlike standard BLS, the retargeted labels ($\bm{T}_i$) can be flexibly balanced between strict zero-one targets ($\bm{Y}_i$) and regression results ($\bm{w}^t$), leading to better classification results. In addition, normal samples can receive higher weights to increase their contributions, while lower weights are assigned to suspicious outliers to reduce their negative effects \cite{chu2019weighted}. Finally, similar to $\bm{Y}_i$, the retargeted label can reformulate as $\bm{T}_i=\left [t_i^1,t_i^2,\dots,t_i^K\right] $ where $t_i^k$ $(k=1,2,\dots,K)$ denotes a class label rather than the real-valued $y_i^k$. Then we can obtain the retargeted label $\bm{T}$ as follows:

\begin{equation} \label{eq:t}
\bm{T}=\begin{bmatrix}
 t_1^1 & \dots  & t_1^K\\
 \vdots & \ddots &\vdots \\\ 
 t_i^1 & \dots & t_i^K
\end{bmatrix} 
\end{equation}

\subsection{Scalable Multi-instance Probabilistic Regression (SMIPR)} \label{ssec:SMIPR}

\begin{figure}[!ht]
\centering
\includegraphics[width=3.5in]{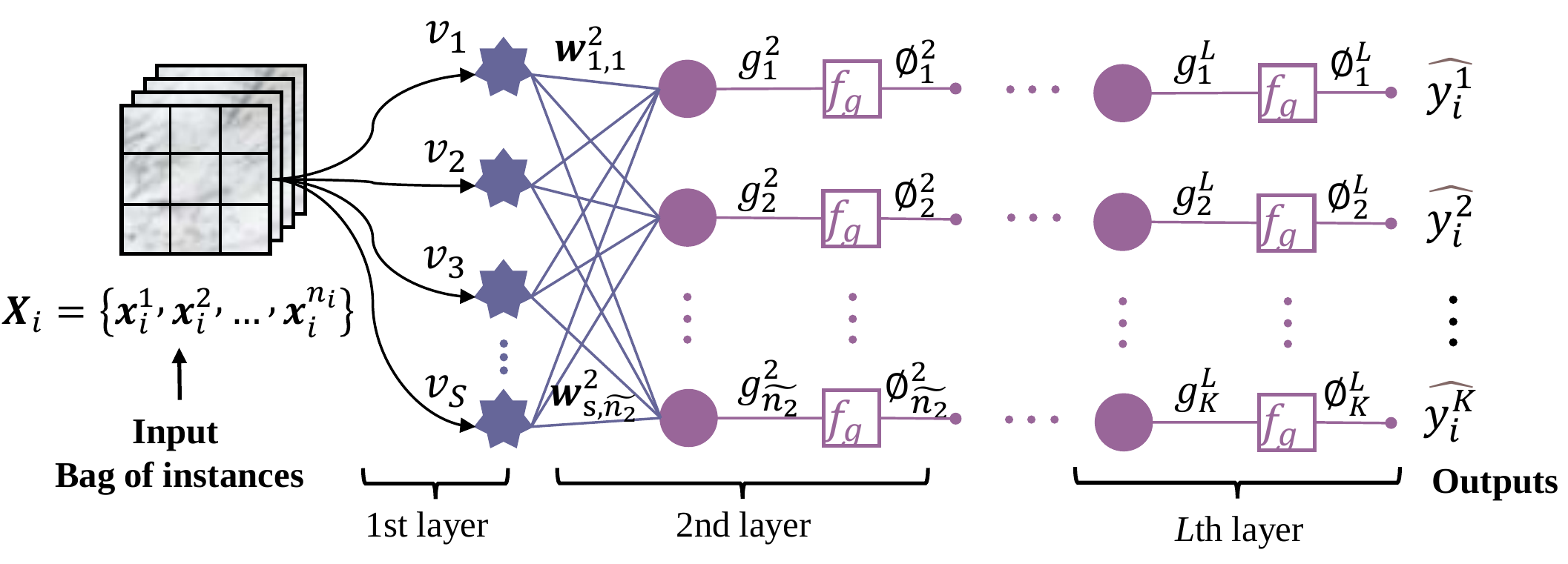}
\caption{Scalable multi-instance probabilistic regressions (SMIPR) structure.}
\label{fig:SMIPR}
\end{figure}

The MIML regression task is the natural extension of traditional (\textit{single instance or single label}) regression to the MIML setting. MIML regression models the sample in the same way as MIML classification, with the important difference that each bag is relevant to several \textit{real-valued} outcomes but not categorical classes. However, each instance in the bag makes a (possibly different) contribution to the bag label \cite{2016book}. For this reason, it becomes necessary to make full use of the probabilistic correlations of different instances in the bag instead of a single score-maximizing instance in predicting the object label. According to the definition about the class-conditional probability density and the prior probability, we can formulate the probability of the joint distribution at the instance-level as below:

\begin{equation} \label{eq:pxy}
P\left (\bm{x}_i^j,\hat{y}_i^c\right )=P\left (\bm{x}_i^j\right )P\left (\hat{y}_i^c\mid\bm{x}_i^j\right) 
\end{equation}
where $\bm{x}_i^j$ indicates the \textit{j}th instance in the \textit{i}th bag while $\hat{y}_i^c$ indicates the \textit{c}th class probability of the ith bag and $\hat{y}_i^c\in \hat{\bm{Y}}_i =\left [\hat{y}_i^c \right ]_{1\times K}$, $c=1,2,...,K$. Note that the index \textit{c} stands for the most probable class, and \textit{K} equals to the number of the correct classes (ground truth). According to the MIML property, the bag includes a series of instances corresponding to \textit{K} possible classes and therefore we have

\begin{equation} \label{eq:pXY}
P\left (\hat{\bm{Y}}_i\mid \bm{X}_i \right ) = {\textstyle \prod_{c=1}^{K}} P\left (\hat{y}_i^c\mid\bm{x}_i^j \right ) 
\end{equation}
where $\bm{X}_i$ indicates the \textit{i}th bag while $\hat{\bm{Y}}_i= \left[\hat{y}_i^1,\hat{y}_i^2,...,\hat{y}_i^K\right]^T$ indicates the K-dimensional output vector. Since Eq. (\ref{eq:pXY}) is computationally intractable, a specifically designed MIML probabilistic regression function g is designed to solve Eq. (\ref{eq:pXY}). The function g of an input bag $\bm{X}_i$ on each of the output vector of the possible label $\bm{Y}_i$ is illustrated in Fig. \ref{fig:SMIPR}. Inspired by minimum squared error criteria \cite{ney1995probabilistic}, no matter whether there is any interdependence between the values of $g\left ( \bm{X}_i,\bm{Y}_i\right )$ for different values of \textit{c}, the squared error attains its absolute minimum if the probabilistic regression function $g\left ( \bm{X}_i,\bm{Y}_i\right )$i) is identical to the class probability $P\left (\hat{\bm{Y}}_i\mid \bm{X}_i \right )$:

\begin{equation} \label{eq:gxy}
g\left ( \bm{X}_i,\bm{Y}_i\right )=P\left (\hat{\bm{Y}}_i\mid \bm{X}_i \right )
\end{equation}

Fig. \ref{fig:SMIPR} shows the scalable multi-instance probabilistic regression (SMIPR) structure employed by BMIML. The regression problem is to determine the $\bm{W}_{PR}$ from training set $\left \{ (\bm{X}_i,\bm{T}_i )\right \}_{i=1}^n$. We define $\bm{W}_{PR} =\left[\bm{w}_{p,q}^l\right]$, and $\bm{w}_{p,q}^l$ indicates the weight connecting the \textit{p}th node in \textit{(l-1)}th layer and the \textit{q}th node in \textit{l}th layer. For the probabilistic regression structure, by regarding each bag as an individual object, the training set $\left \{ \bm{X}_i\right \}_{i=1}^n$ is clustered in the first layer \textit{(l=1)} into \textit{S disjoint groups} of bags $\left \{ G_p\right \}_{p=1}^S\left ( G_{s1}\cap_{s1\neq s2}G_{s2}=\emptyset \right )$ with ${\textstyle\bigcup_{p=1}^{S}} G_p =\left\{ \bm{X}_i\right \}_{i=1}^n$ by \textit{k-means algorithm} \cite{yuan2019research}. After the clustering process, the training set is divided into \textit{S} partitions and their medoids $v_p$ are decided as:

\begin{equation} \label{eq:vp}
v_p = arg\min_{\bm{A}\in G_p}{\textstyle \sum_{\bm{B}\in G_p}} dist\left (\bm{A},\bm{B}\right ) 
\end{equation}
where $dist\left (\bm{A},\bm{B}\right )$ denotes the \textit{Hausdorff distance} \cite{karimi2019reducing} between two bags of instances
$\bm{A}=\left \{\bm{a}_1,\bm{a}_2,\dots,\bm{a}_{N1}\right \}$ and $\bm{B}=\left \{\bm{b}_1,\bm{b}_2,\dots,\bm{b}_{N2}\right \}$, which can be defined as: $dist\left (\bm{A},\bm{B}\right)=\max\left \{ \max_{\bm{a}\in \bm{A}} \min_{\bm{b}\in \bm{B}} \left \| \bm{a}-\bm{b} \right \|,\max_{\bm{b}\in \bm{B}} \min_{\bm{a}\in \bm{A}} \left \| \bm{b}-\bm{a} \right \|\right \}$ where $ \left \| \bm{a}-\bm{b} \right \|$ measures the distance between instances $\bm{a}$ and $\bm{b}$ When the number of the layers is set to 2 (i.e., \textit{l}=2), the numbers of input and output nodes are fixed so that $\bm{W}_{PR} =\left[\bm{w}_{p,q}^l\right]_{S\times K}$ where \textit{S} is the maximum number of the clusters (input), and \textit{K} is the maximum number of output classes. For the \textit{l}th ${2< l< L}$ layer, the maximum number of nodes is defined as $\tilde{n}_l$ while the number of output nodes (\textit{l=L}) is still set to \textit{K}, which is the scalable part of the multi-instance probabilistic regression structure, as shown in Fig. \ref{fig:SMIPR}. Then the weights $\left[\bm{w}_{p,q}^l\right]$ can be optimized by minimizing the following sum-of-squares error function:

\begin{equation} \label{eq:e}
E=\frac{1}{2} {\textstyle \sum_{i=1}^{n}} {\textstyle \sum_{q=1}^{K}}\left \{g_{q}^{l}(\bm{X}_i)-t_{i}^{q}\right \}^2
\end{equation}
where $t_i^q$ is the desired output values in output layer (\textit{l = L} and \textit{q = 1} to \textit{K}) of $\bm{X}_i$ on the \textit{q}th class with the elements $\left[t_i^q\right]_{n\times K= \bm{T}}$, and $g_{q}^{l}(\bm{X}_i)$ is defined as:

\begin{equation} \label{eq:gql}
g_{q}^{l}(\bm{X}_i)=\begin{cases}
   {\textstyle \sum_{p=1}^{\tilde{n}_{l-1}}}\bm{w}_{p,q}^l\phi_p^{l-1}(\bm{X}_i) & \text{ if } l>2  \\
  {\textstyle \sum_{p=1}^{S}}\bm{w}_{p,q}^l\phi_p^{l}(\bm{X}_i)& \text{ if } l=2
\end{cases}
\end{equation}
where \textit{p} and \textit{q} are the numbers of nodes in the (\textit{l-1})th and \textit{l} layer, respectively. Finally, $\phi_p^{l}(\bm{X}_i)$ can be calculated as below:

\begin{equation} \label{eq:qql}
\phi_p^{l}(\bm{X}_i)=\begin{cases}
  {\textstyle \sum_{p=1}^{\tilde{n}_{l-1}}}\bm{w}_{p,q}^l f_g(\phi_p^{l-1}(\bm{X}_i))& \text{ if } l>2 \\
  dist(\bm{X}_i,v_p)& \text{ if } l=2
\end{cases}
\end{equation}
where $f_g(\cdot)$ is an activation function (e.g., \textit{sigmoid}) and the weights $\bm{w}_{p,q}^l=\left\{\begin{matrix}
  0,&p=q \\
  1,&p\neq q
\end{matrix}\right.$ if \textit{l}=2, $\bm{w}_{p,q}^l,(l>2)$ can be updated using gradient descent:
\begin{equation} \label{eq:wpq}
\bm{w}_{p,q}^{l+1}= \bm{w}_{p,q}^{l}+\eta(f_g(\phi_p^{l}(\bm{X}_i))\bigtriangleup^l)
\end{equation}
where $\eta$ is a learning rate and $\bigtriangleup^l$ is denoted as the gradient of the \textit{l}th layer obtained in back-propagation:

\begin{equation} \label{eq:ll}
\bigtriangleup^l=\begin{cases}
  \bigtriangleup^{l+1} \bm{w}^{l+1}\bm{F}^l,& \text{ if } 2< l< L \\
  (\bm{T}-g(\bm{X}))^T\bm{F}^l& \text{ if } l=L
\end{cases}
\end{equation}
where $\bm{F}^l=\begin{bmatrix}
  f_g^{'}(\phi_{i,1}^{l})&\cdots  &  0      &\cdots  & 0\\
  \vdots                 &\ddots   &\vdots  &\cdots &\vdots \\
  0                      &\cdots  & f_g^{'}(\phi_{i,p}^{l}) &\cdots  &0 \\
  \vdots   &\cdots  &\vdots  & \cdots &\vdots \\
  0&  \cdots  &0  &\cdots  &f_g^{'}(\phi_{\tilde{n}_{l},1}^{l})
\end{bmatrix}$.

\subsection{Interactive Decision Optimization (IDO)} \label{ssec:ido}

To combine the classification result from AWLEL and probabilistic regression from SMIPR, an interactive module called IDO is designed, which forms an end-to-end learning framework to reduce user intervention (individual learning of bags and instances) and achieve better classification results. By combining Eqs. (\ref{eq:ti}), (\ref{eq:rwt}) (\ref{eq:pXY}) and (\ref{eq:e}), the predicted label of a bag $\bm{X}_i$ can be obtained as follows:

\begin{equation} \label{eq:fd}
\begin{aligned}
&\hat{\bm{Y}}_i  = \left \{f_{decision}^c(g_k^l(\bm{X}_i))\right \}_{c = 1}^K \\
& s.t. g_k^l(\bm{X}_i)\in \left [\min (\bm{T}_1),\max (\bm{T}_1)\right ]
\end{aligned}
\end{equation}
where $f_{decision}^c$ is the decision function for \textit{c}th class $({c=1,2,\dots,K})$, and it can be formulated as 

\begin{equation} \label{eq:fdc}
f_{decision}^c(r)=\left\{\begin{matrix}
  1,&\rho (r)> \tau  \\
  0,&\rho (r)\le  \tau 
\end{matrix}\right.
\end{equation}
where $\tau$ is the user-defined decision threshold, and it is individually set for every \textit{c}, and $\rho(\cdot)$ represents the softmax function. In our experiment, we set $\tau=0.8$, that is, only when the probability of belonging to class $c$ is larger than 0.8, it can be classified as class $c$.

\subsection{Optimization Strategy} \label{ssec:os}

In this section, we give the optimal solution of Eq. (\ref{eq:rwt}) through the strategy of the ADMM algorithm \cite{wahlberg2012admm}. For simplicity, Eq. (\ref{eq:rwt}) is reformulated with the \textit{Lagrangian function} as

\begin{equation} \label{eq:fw}
f_L(\bm{w}^t)=\left \| \sqrt{\bm{\Gamma}} (\bm{A}\bm{w}_t-\bm{T}) \right \|_2^2 +\lambda  \left \| \bm{w}_t \right \|_2^2
\end{equation}
\begin{equation} \label{eq:ft}
f_L(\bm{T})= \left \| \sqrt{\bm{\Gamma}} (\bm{A}\bm{w}_t)-\bm{T} \right \|_2^2+\vartheta\left \| \sqrt{\bm{\Omega} } (\bm{T}-\bm{Y}) \right \|_2^2
\end{equation}
Fix $\bm{T}$ Update $\bm{w}^t$: When $\bm{T}$ are known, taking the derivative of Eq. (\ref{eq:fw}) and setting it to 0. Then Eq. (\ref{eq:fw}) can be written as the following optimization with respect to $\bm{w}^t$:

\begin{equation} \label{eq:27}
\begin{aligned}
&2\bm{A}^T\bm{\Gamma}(\bm{A}\bm{w}_t-\bm{T})+2\lambda \bm{w}_t=0\\
&\Rightarrow \bm{w}_t= (\lambda\bm{I}+\bm{A}^T\bm{\Gamma}\bm{A})^{-1}\bm{A}^T\bm{\Gamma}\bm{T}
\end{aligned}
\end{equation}
Fix $\bm{w}^t$ Update $\bm{T}$: Since $\bm{w}^t$ is fixed, similarly setting the derivative of Eq. (\ref{eq:fw}) to 0, we arrive at

\begin{equation} \label{eq:tt}
\begin{aligned}
&2\bm{\Gamma}(\bm{A}\bm{w}_t-\bm{T})+2\vartheta\bm{\Omega}(\bm{T}-\bm{Y})=0\\
&\Rightarrow \bm{T}= (\bm{\Gamma}+\vartheta\bm{\Omega})^{-1}(\bm{\Gamma}\bm{A}\bm{w}_t+\vartheta\bm{\Omega}\bm{Y})
\end{aligned}
\end{equation}

Based on the above results, we alternatively update $\bm{T}$ and $\bm{w}_t$ through the Eqs. (\ref{eq:wt}) and (\ref{eq:tt}) until convergence or the termination condition is satisfied.

\begin{algorithm}[H]
\caption{BMIML}\label{alg:alg1}
\begin{algorithmic}
\STATE 
\STATE \textbf{Input:} The matrix representation of $i^{th}$ samples in all $n$ training samples: $\mathbf{X}_i$; the set of training instances matrix (bags): $\left \{ \mathbf{x}_i^j \right \}_{j=1}^{n_i}$; the label matrix for all n training samples and bags: $Y_i, i=1$ to $n$, decision threshold $\tau$.
\STATE \textbf{Output:} Predicted Label $\hat{\mathbf{Y}}_i$.
\STATE \textbf{Steps of label enhancement learning:} 
\STATE \hspace{0.5cm} Calculate $Z$ and $H$ in the board learning system with the input $X$ according to Eq.(1)-Eq.(3);
\STATE \hspace{0.5cm} Calculate $w^t$ and $T$ by solving the problem of Eq. (11) using Eq.(26)-Eq.(27);
\STATE \textbf{Steps of multi-instance probabilistic regression:}
\STATE \hspace{0.5cm} \textbf{Do}
\STATE \hspace{1cm} \textbf{For} $t = 1 \ to \ n$
\STATE \hspace{1.5cm} Generate distance matrix of $X_i$ according to Eq.(16);
\STATE \hspace{1.5cm} Clustering instances in $X_i$ into $S$ clusters:\\
\hspace{3cm} $ {\textstyle \bigcup_{p=1}^{S}} G_p=\left \{ \mathbf{X}_i \right \} ^n_{i=1}$;
\STATE \hspace{1.5cm} Update the weights of probabilistic regression $W_{PR}$ according to Eq.(20);
\STATE \hspace{1cm} \textbf{END}
\STATE \hspace{0.5cm} \textbf{until} \textit{Convergence}
\STATE \textbf{Classification:}
\STATE \hspace{0.5cm} Predict the label according to Eq.(22):\\ \hspace{2.5cm} $\hat{\mathbf{Y}}_i \gets f_{decision}^c (g_L^k (\mathbf{X}_i))$
\end{algorithmic}
\label{algorithm:BMIML}
\end{algorithm}

\section{Experimental} \label{sec:exp}

\subsection{Datasets} \label{ssec:data}

\begin{table}[!ht]
\caption{Properties of datasets\label{tab:table2}}
\centering
\begin{tabular}{llllll}
\hline
\multicolumn{1}{c}{dataset} & \multicolumn{1}{c}{Instances} & Bags    & \multicolumn{1}{c}{Labels} & \multicolumn{1}{c}{Image Type}                           & \multicolumn{1}{c}{Resolution} \\ \hline
\textit{NuCLS}              & 5,432                         & 1,358   & 7                          & WSI                                                      & Various                        \\
\textit{Breast}               & 2,416                         & 151     & 22                         & WSI                                                      & 1024*1024                      \\
\textit{Pannuke}            & 31,616                        & 7,904   & 5                          & WSI                                                      & 256*256                        \\
\textit{ODR}                & 90,000                        & 10,000  & 8                          & \begin{tabular}[c]{@{}l@{}}fundus \\ photos\end{tabular} & 576*576                        \\
\textit{NIH}                & 896,960                       & 112,120 & 14                         & X-ray                                                    & 512*512                        \\ \hline
\end{tabular}
\end{table}

\begin{table*}[!b]
\caption{Comparison results (mean $\pm$ std.) on three medium data sets\label{tab:table3}}
\centering
\begin{tabular}{lcccccccc}
\hline
Methods & MIMLNN          & MIMLSVM            & MIMLmiSVM       & MIMLkNN   & MIMLBoost          & MIMLfast  & DeepMIML & \textbf{BMIML}              \\ \hline
\multicolumn{9}{l}{\textit{NuCLS}}                                                                                                            \\ \hline
H.L.↓     & .125$\pm$.004       & .106$\pm$.008          & .494$\pm$.017       & .233$\pm$.005 & \underline {.116$\pm$.025}    & .253$\pm$.028 &.202$\pm$.030          & \textbf{.088$\pm$.030} \\
O.E.↓     & .264$\pm$.010       & .132$\pm$.027          & .136$\pm$.043       & .284$\pm$.022 & \textbf{.029$\pm$.001} & .583$\pm$.061 &.525$\pm$.008          & \underline{.037$\pm$.015}    \\
R.L.↓     & .077$\pm$.002       & \textbf{.041$\pm$.020} & .368$\pm$.017       & .380$\pm$.023 & .099$\pm$.005          & .392$\pm$.004 &.325$\pm$.019          & \underline{.043$\pm$.010}    \\
A.P.↑     & .857$\pm$.041       & \underline{.941$\pm$.006}    & .856$\pm$.028       & .757$\pm$.007 & .921$\pm$.009          & .722$\pm$.011 & .815$\pm$.046         & \textbf{.968$\pm$.007} \\ \hline
\multicolumn{9}{l}{\textit{Breast}}                                                                                                           \\ \hline
H.L.↓     & \underline {.293$\pm$.060} & .297$\pm$.011          & .511$\pm$.041       & .297$\pm$.033 & .460$\pm$.030          & .318$\pm$.021 &.541$\pm$.032         & \textbf{.290$\pm$.017} \\
O.E.↓     & .219$\pm$.013       & .206$\pm$.032          & .183$\pm$.003       & .250$\pm$.062 & \textbf{.013$\pm$.001} & .500$\pm$.016 &.500$\pm$.003         & \underline {.094$\pm$.001}    \\
R.L.↓     & {\underline .204$\pm$.007} & .196$\pm$.050          & .438$\pm$.028       & .483$\pm$.010 & .943$\pm$.041          & .493$\pm$.022 &.502$\pm$.046          & \textbf{.172$\pm$.004} \\
A.P.↑     & .822$\pm$.028       & \underline {.832$\pm$.064}    & .770$\pm$.071       & .599$\pm$.025 & .624$\pm$.019          & .591$\pm$.016 &.530$\pm$.026          & \textbf{.854$\pm$.021} \\ \hline
\multicolumn{9}{l}{\textit{Pannuke}}                                                                                                          \\ \hline
H.L.↓     & .299$\pm$.036       & \underline {.285$\pm$.041}    & .510$\pm$.018       & .299$\pm$.005 & N/A                & .377$\pm$.011 & N/A      & \textbf{.276$\pm$.005} \\
O.E.↓     & .250$\pm$.012       & \textbf{.167$\pm$.024} & \underline{.182$\pm$.033} & .200$\pm$.022 & N/A                & .600$\pm$.032 & N/A      & .212$\pm$.038          \\
R.L.↓     & .209$\pm$.030       & \underline {.189$\pm$.006}    & .438$\pm$.009       & .509$\pm$.036 & N/A                & .465$\pm$.031 & N/A      & \textbf{.151$\pm$.014} \\
A.P.↑     & .806$\pm$.042       & \underline {.823$\pm$.045}    & .770$\pm$.013       & .441$\pm$.040 & N/A                & .439$\pm$.060 & N/A      & \textbf{.846$\pm$.003} \\ \hline
\multicolumn{9}{l}{\textit{$\uparrow$($\downarrow$) indicates that the larger (smaller) the value, the better the performance;Bold indicates the best performance of this metric;} }\\
\multicolumn{9}{l}{\textit{\underline{underline} indicates the next best performance of this metric; N/A represents that no result was obtained in 72 hours}.} \\
\end{tabular}
\end{table*}

\begin{table*}[!b]
\caption{Classification average precision (AP) (mean $\pm$ std.) of comparison algorithms on two large data sets with various data sizes\label{tab:table4}}
\centering
\begin{tabular}{cccccccccc}
\hline
\multicolumn{2}{c}{Dataset (Size)}     & MIMLNN     & MIMLSVM   & MIMLmiSVM & MIMLkNN   & MIMLBoost & MIMLfast   & DeepMIML & \textbf{BMIML}               \\ \hline
\multirow{5}{*}{\textit{ODR}} & \#2K   & .670$\pm$.080  & .649$\pm$.002 & .700$\pm$.088 & .214$\pm$.022 & .580$\pm$.088 & .465$\pm$.070  & \underline{.686$\pm$.002}         & \textbf{.727$\pm$.056}          \\
                              & \#4K   & .741$\pm$.047  & \underline{.747$\pm$.010} & N/A       & .225$\pm$.060 & .604$\pm$.036 & .466.$\pm$.048 &N/A          & \textbf{.778$\pm$.028} \\
                              & \#6K   & .756$\pm$.020  & \underline{.775$\pm$.003} & N/A       & .243$\pm$.031 & N/A       & .483$\pm$.005  &N/A          & \textbf{.835$\pm$.039} \\
                              & \#8K   & .778$\pm$.031  & \underline{.797$\pm$.014} & N/A       & .294$\pm$.090 & N/A       & .506.$\pm$.043 &N/A          & \textbf{.878$\pm$.047} \\
                              & \#10K  & .794$\pm$.018  & \underline{.846$\pm$.041} & N/A       & .342$\pm$.066 & N/A       & .512$\pm$.056  &N/A          & \textbf{.917$\pm$.030} \\ \hline
\multirow{4}{*}{\textit{NIH}} & \#30K  & .391$\pm$.090  & \underline{.508$\pm$.080} & N/A       & .271$\pm$.082 & N/A       & .344$\pm$.026  &N/A          & \textbf{.536$\pm$.046} \\
                              & \#60K  & .396$\pm$.002  & \underline{.511$\pm$.052} & N/A       & .274$\pm$.091 & N/A       & .350$\pm$.075  &N/A          & \textbf{.574$\pm$.002} \\
                              & \#90K  & .396$\pm$.0081 & \underline{.519$\pm$.019} & N/A       & .274$\pm$.026 & N/A       & .370$\pm$.036  &N/A          & \textbf{.603$\pm$.041} \\
                              & \#120K & .397$\pm$.041  & \underline{.527$\pm$.066} & N/A       & N/A       & N/A       & .376$\pm$.028  &N/A          & \textbf{.661$\pm$.039} \\ \hline
\multicolumn{10}{l}{\textit{N/A means that no result was obtained in 72 hours}.}\\
\multicolumn{10}{l}{\textit{\textbf{Bold} indicates the best performance of this metric; \underline{underline} indicates the next best performance of this metric}.} \\
\end{tabular}
\end{table*}

Our experiment was conducted over 5 real-world data sets from TCGA and Github for multi-label medical image classification about \textit{whole-slide images} (WSIs), \textit{X-ray}, and \textit{computed tomography }(CT), etc. The \textbf{NuCLS data set} \cite{amgad2021nucls} is collected by TCGA, which contains 1358 WSIs for breast cancer with 7 possible labels. The \textbf{Breast Cancer Semantic Segmentation data set} (BCSS) \cite{amgad2019structured} consists of 151 hematoxylin and eosin stained WSIs corresponding to 22 histologically-confirmed breast cancer cases. \textbf{Pannuke data set} \cite{gamper2020pannuke} consists of 7904 WSIs across 19 different tissue types with 5 possible labels. The \textbf{ODR data set} \cite{li2020benchmark} contains 10,000 \textit{color retinal fundus} images annotated with 8 possible labels. \textit{ODR} is collected by Shanggong Medical Technology Co., Ltd. from different hospitals/medical centers in China. In these institutions, fundus images are captured by various cameras in the market, such as Canon, Zeiss and Kowa, under various image resolutions. The largest data set \textbf{NIH Cheat X-ray data set} collected by \textit{the NClinical Center} (clinicalcenter.nih.gov) and \textit{National Library of Medicine} (www.nlm.nih.gov) contains 112,120 images with 14 possible labels, and each image is represented with a bag of 4 instances. The properties of these data sets are summarized in Table \ref{tab:table2}. For each data set, 60\% of the data are randomly selected for training, 10\% for validation, and the remaining 30\% for testing. In our experiment, the results are recorded after 10 epochs of model training where the instances in the bags were shuffled in each epoch. 

\subsection{Settings} \label{ssec:settings}
To verify the advantage of BMIML on the task of multi-label medical image classification, seven state-of-the-art MIL approaches were compared: MIMLNN \cite{zhou2012multi}, MIMLSVM \cite{zhou2006multi}, MIMLmiSVM \cite{zhou2012multi}, MIMLkNN \cite{zhang2010k}, MIMLBOOST \cite{zhou2006multi}, MIMLfast \cite{huang2018fast}, DeepMIML \cite{feng2017deep}. For fair comparison, the parameters of all the compared approaches are determined in the same way if no value is suggested in their literature. Instances are simply divided according to the size of the original image. In our experiment, we try to ensure that the size of each instance is about 64 * 64. Thus, the number of instances in each bag is equal to the resolution of the original image divided by 64 (See Table \ref{tab:table2} for details of the data sets). Of course, other methods can also be used to generate the instances. Four commonly used MIML metrices are employed for performance evaluation: \textit{hamming loss} (HL), \textit{one error} (OE), \textit{ranking loss} (RL), and \textit{average precision} (AP). All definitions of these metrices can be found in [22, 40]. For better performance evaluation, 10-fold cross validation is conducted on a machine with i7-9700k 3.60GHz CPU and 32 GB RAM memory.

\subsection{Performance Comparison} \label{ssec:performance}
\textbf{Medium data sets}
The comparison results on three medium data sets are listed in Table \ref{tab:table3}. BMIML achieves the best performance in most cases, MIMLNN and MIMLkNN work steadily on all the data sets but are not competitive when compared with BMIML. Although MIMLSVM achieves comparable results with our proposed methods in some cases, it is less effective on large data sets in Table \ref{tab:table4}. MIMLBoost and DeepMIML can handle only two smallest data sets (\textit{NuCL}S and \textit{Breast}), and do not yield very good performance. MIMLfast works very poorly over all metrics on these three data sets. When the number of instances increases, its accuracy drops obviously. 

\begin{figure*}[!t]
\centering
\subfloat[]{\includegraphics[width=2in]{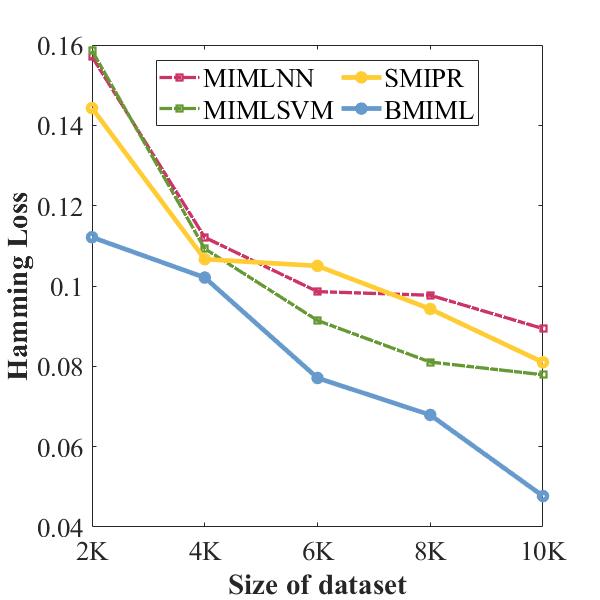}%
\label{fig:odr_hl}}
\subfloat[]{\includegraphics[width=2in]{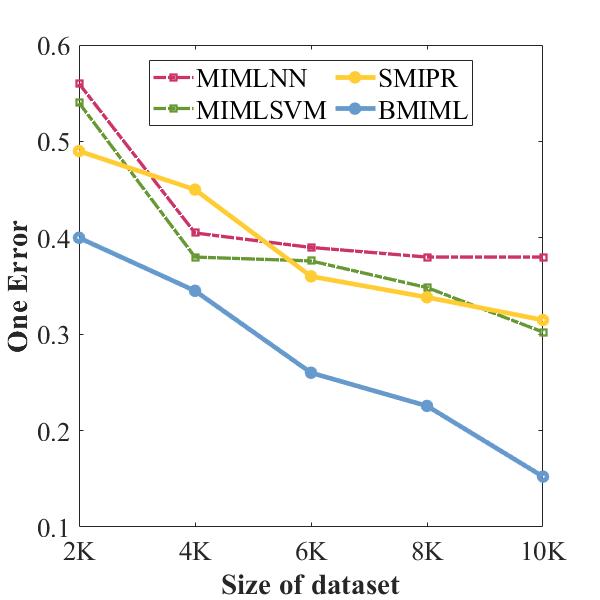}%
\label{fig:odr_oe}}
\subfloat[]{\includegraphics[width=2in]{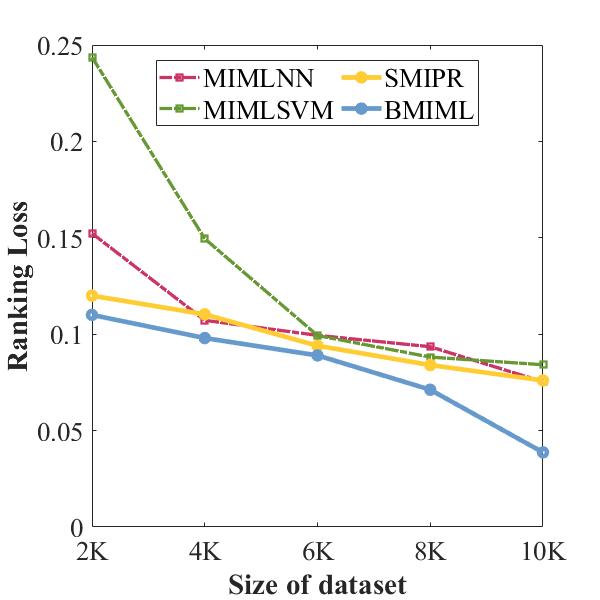}%
\label{fig:odr_rl}}
\caption{Comparison results on \textit{ODR} with varying data size; the values smaller, the performance better.}
\label{fig:ODR_result}
\end{figure*}

\begin{figure*}[!t]
\centering
\subfloat[]{\includegraphics[width=2in]{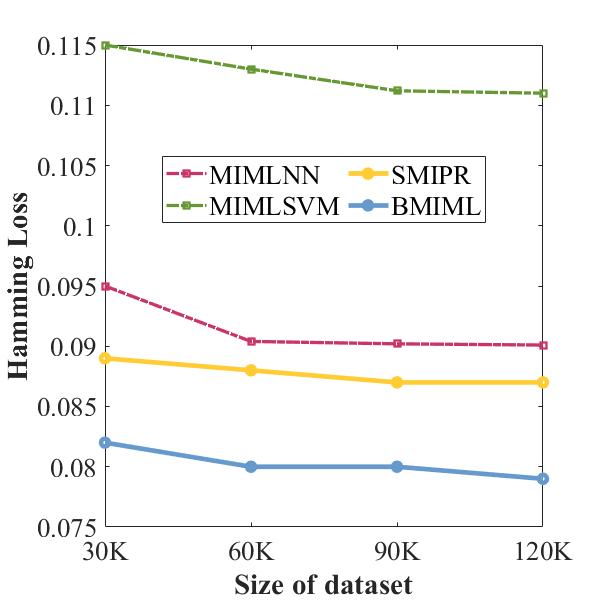}%
\label{fig:nih_hl}}
\subfloat[]{\includegraphics[width=2in]{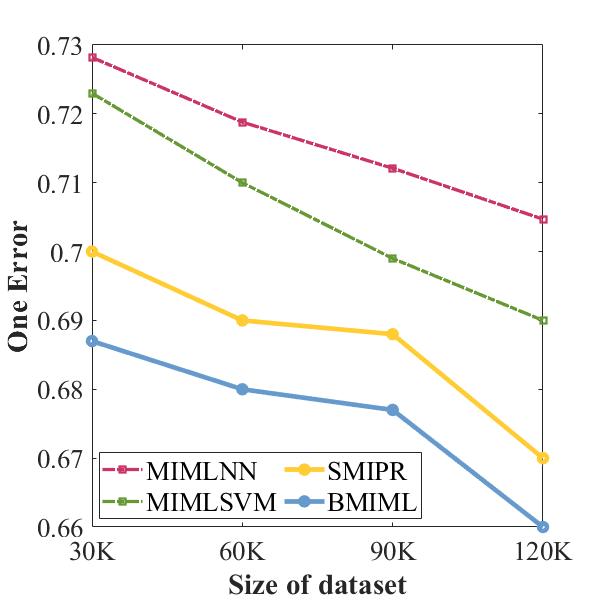}%
\label{fig:nih_oe}}
\subfloat[]{\includegraphics[width=2in]{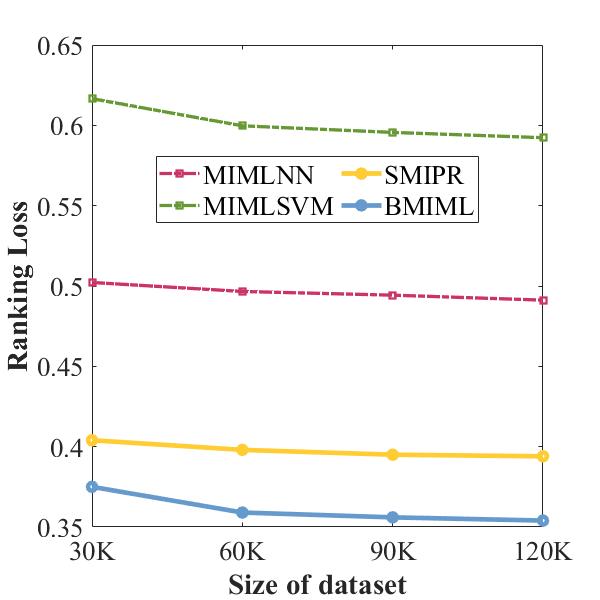}%
\label{fig:nih_rl}}
\caption{Comparison results on \textit{NIH} with varying data size; the values smaller, the performance better.}
\label{fig:NIH_result}
\end{figure*}

\textbf{Large data sets}
\textit{ODR} and \textit{NIH} contain 10,000 and 112,120 bags respectively, which are too large for most existing MIML approaches. Therefore, the comparison was conducted on their subsets with various data sizes. The number of bags in \textit{ODR} ranges from 2,000 to 10,000, and the number of bags in \textit{NIH} ranges from 30,000 to 120,000, and the average precision (AP) is shown in Table \ref{tab:table4}. For \textit{NIH}, MIMLmiSVM and MIMIBoost cannot return any result after 72 hours even for the smallest data size (30K). Similarly, in \textit{ODR}, MIMIBoost can only handle up to 4,000 bags while MIMLmiSVM up to 2,000 bags. In Table \ref{tab:table4}, the AP performance of MIMLkNN and MIMLfast are not comparable with other methods. For this reason, their performances on HL, OE and RL are not shown in Figs. \ref{fig:ODR_result} and \ref{fig:NIH_result}. In Figs. \ref{fig:ODR_result} and \ref{fig:NIH_result}, the trends of HL, OE, and RL drop along with increasing data sizes on the two large data sets \textit{ODR} and \textit{NIH}, respectively, while our proposed BMIML is obviously better than the others. Furthermore, BMIML is much more stable and effective than other methods on \textit{NIH} data set for four evaluation the metrics (HL, OE, RL, AP). 

\begin{table*}[!ht]
\caption{Classification performance (mean $\pm$ std.) of AWLEL, SMIPR, and BMIML on three medium data sets\label{tab:table5}}
\centering
\begin{tabular}{lcccllll}
\hline
Datasets                 & AWLEL                  & SMIPR                  & IDO                    & \multicolumn{1}{c}{H.L.↓} & \multicolumn{1}{c}{O.E.↓} & \multicolumn{1}{c}{R.L.↓} & \multicolumn{1}{c}{A.P.↑} \\ \hline
\multirow{3}{*}{\textit{NuCLS}}   & $\surd$ &                        &                        & .463$\pm$.021            & .250$\pm$.011            & .169±.042                 & .736±.080                \\
                         &                        & $\surd$ &                        & .105±.005                & .056±.020                & .051±.003                & .908±.012                \\
                         & $\surd$ & $\surd$ & $\surd$ & \textbf{.088±.030}                & \textbf{.037±.015}                & \textbf{.043±.010}                & \textbf{.968±.00}7                \\ \hline
\multirow{3}{*}{\textit{Breast}}  & $\surd$ &                        &                        & .600±.020                & .500±.013                & .543±.050                & .548±.016                \\
                         &                        & $\surd$ &                        & .291±.040                & .193±.003                & .190±.006                & .833±.019                \\
                         & $\surd$ & $\surd$ & $\surd$ & \textbf{.290±.017}                & \textbf{.094±.001}                & \textbf{.172±.004}                & \textbf{.854±.021}                \\ \hline
\multirow{3}{*}{\textit{Pannuke}} &$\surd$ &                        &                        & .617±.070                & .400±.023                & .594±.061                & .432±.084                \\
                         &                        &$\surd$ &                        & .290±.031                & .238±.022                & .187±.030                & .825±.051                \\
                         &$\surd$ & $\surd$ & $\surd$ & \textbf{.276±.005}                & \textbf{.212±.038}                & \textbf{.151±.014}                & \textbf{.846±.003}                \\ \hline
\end{tabular}
\end{table*}

\begin{table}[!ht]
\caption{Classification average precision (AP) (mean $\pm$ std.) of AWLEL, SMIPR and BMIML on two large datasets with various data sizes \label{tab:table6}}
\centering
\begin{tabular}{ccccc}
\hline
\multicolumn{2}{l}{datasets (size)}    & AWLEL               & SMIPR               & BMIML                        \\ \hline
\multirow{5}{*}{\textit{ODR}} & \#2K   & .462$\pm$.064          & .714$\pm$.008          & \textbf{.727$\pm$.056}          \\
                              & \#4K   & .469$\pm$.025          & .755$\pm$.016          & \textbf{.778$\pm$.028}          \\
                              & \#6K   & .502$\pm$.016          & .785$\pm$.054          & \textbf{.835$\pm$.039 }         \\
                              & \#8K   & .508$\pm$.076          & .837$\pm$.023          & \textbf{.878$\pm$.047}          \\
                              & \#10K  & .519$\pm$.033 & .864$\pm$.033 & \textbf{.917$\pm$.030} \\ \hline
\multirow{4}{*}{\textit{NIH}} & \#30K  & .310.$\pm$.025          & .535$\pm$.021          & \textbf{.536$\pm$.046}          \\
                              & \#60K  & .318.$\pm$.030          & .554$\pm$.045          & \textbf{.574$\pm$.002}          \\
                              & \#90K  & .320.$\pm$.011          & .580$\pm$.002          & \textbf{.603$\pm$.041}          \\
                              & \#120K & .331.$\pm$.061          & .612$\pm$.033          & \textbf{.661$\pm$.039}          \\ \hline
\end{tabular}
\end{table}

\begin{table*}[!t]
\caption{Training time comparison (in seconds)\label{tab:table7}}
\centering
\begin{tabular}{lcccccccccc}
\hline
Datasets           & MIMLNN   & MIMLSVM  & MIMLmiSVM & MIMLkNN  & MIMLBoost        & MIMLfast       & DeepMIML  & \textbf{AWLEL} & \textbf{SMIPR} & \textbf{BMIML} \\ \hline
NuCLS              & 63.7     & 189.6    & 13672.8   & 178.4    & 32165.2          & \textbf{14.7} &50980.1    & 15.9           & 73.4           & 102.1          \\
\textit{Breast}    & 213.3    & 832.5    & 130212.2  & 899.38   & 314913.4         & 31.4          &499114.2   & \textbf{26.2}  & 99.3           & 149.94         \\
\textit{Pannuke}   & 10918.4  & 40600.3  & 390637.3  & 42383.2  & N/A              & 972.5         &N/A        & \textbf{550.3} & 9174.7         & 9691.3         \\
ODR\_2K            & 110.5    & 424.5    & 34689.6   & 584.57   & 78909.2 & 49.7          &788860.1   & \textbf{15.8}  & 108.6          & 157.3          \\
\textit{ODR\_4K}   & 592.2    & 1677.0   & N/A       & 1669.3   & 356740.7         & 109.4         &N/A       & \textbf{54.3}  & 600.4          & 685.4          \\
\textit{ODR\_6K}   & 1326.3   & 3848.5   & N/A       & 4248.5   & N/A              & 357.4         &N/A      & \textbf{180.7} & 1510.7         & 1690.7         \\
ODR\_8K            & 4875.1   & 7056.7   & N/A       & 7656.9   & N/A              & 745.9         &N/A       & \textbf{252.6} & 3321.5         & 3651.5         \\
\textit{ODR\_10K}  & 12832.4  & 13446.8  & N/A       & 13680.2  & N/A              & 972.5         &N/A       & \textbf{444.6} & 6872.3         & 7349.1         \\
\textit{NIH\_30K}  & 9454.7   & 10577.6  & N/A       & 10839.8  & N/A              & \textbf{500.9} &N/A   & 1303.6         & 4406.9         & 5352.3         \\
\textit{NIH\_60K}  & 35718.9  & 54054.6  & N/A       & 131671.5 & N/A              & \textbf{1062.5} &N/A  &  3012.7         & 16994.8        & 18657.2        \\
\textit{NIH\_90K}  & 85520.1  & 217528.8 & N/A       & N/A      & N/A              & \textbf{1720.9} &N/A  & 12303.6        & 38416.6        & 47335.3        \\
\textit{NIH\_120K} & 135587.2 & N/A      & N/A       & N/A      & N/A              & \textbf{2420.1} &N/A  & 21077.3        & 53180.8        & 68459.3        \\ \hline
\end{tabular}
\end{table*}

\begin{figure*}[!ht]
\centering
\includegraphics[width=6.5in]{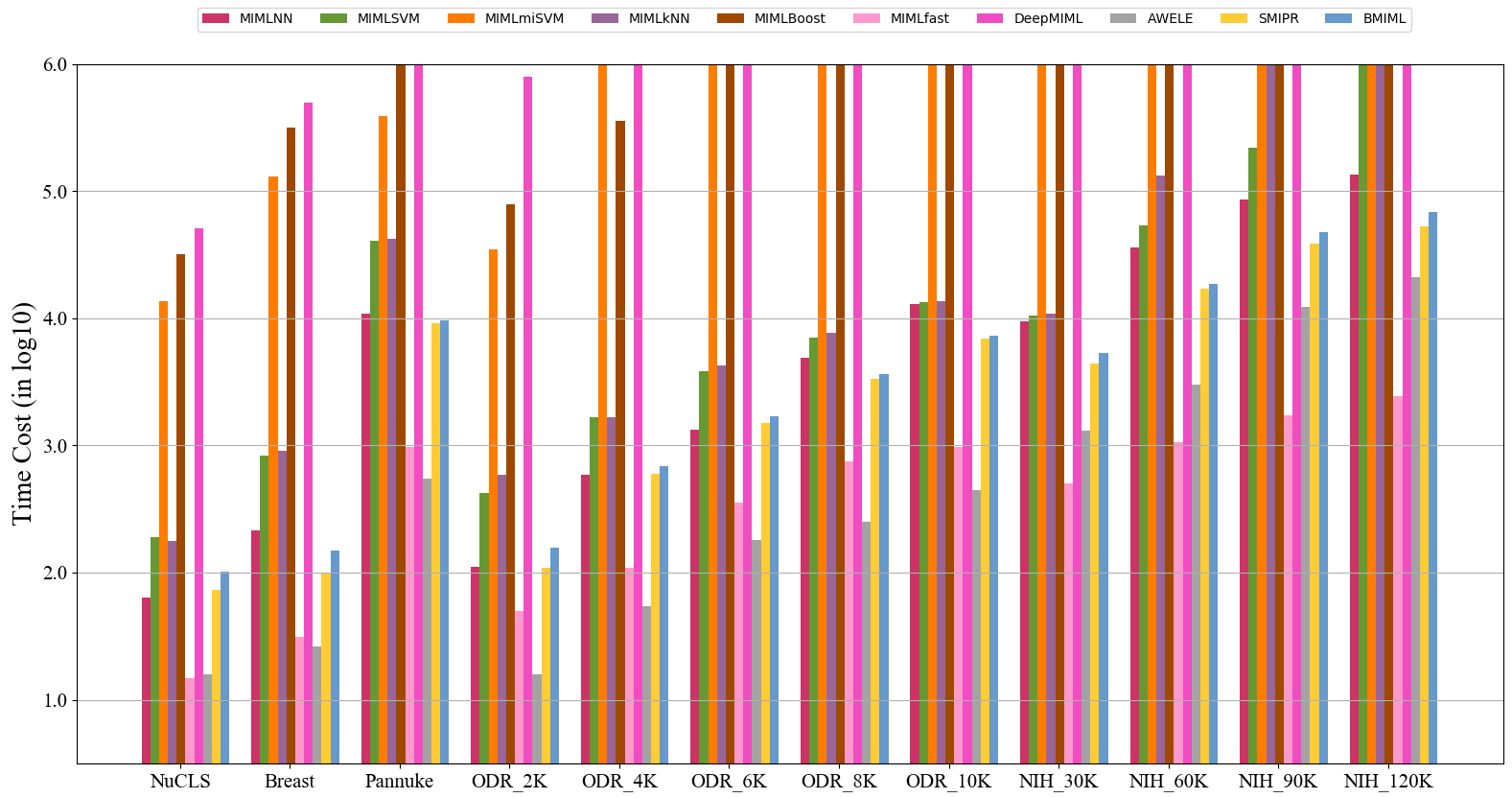}
\caption{Training time comparison (in seconds).}
\label{fig:times}
\end{figure*}

\subsection{Module Analysis} \label{ssec:module}

To evaluate the performance of the two proposed modules (AWLEL and SMIPR) ablation studies are conducted. The number of layers \textit{l} in BMIML and SMIPR is both set to 3. For \textbf{medium datastes}, as shown in Table \ref{tab:table5}, BMIML achieves the best performance and the proposed SMIPR alone performs the next best in most cases which validate our idea to consider both global view and local view rather than local view only. The performance of AWELE alone in various metrics is not competitive to SMIPR and BMIML since the ability of BLS feature learning is relatively weak. As illustrated in Table \ref{tab:table6}, for \textbf{large data sets} AWELE does not work well while SMIPR is relatively better but still not comparable to BMIML. With the increasing data set sizes, the advantage of BMIML becomes more and more obvious. Combined with Tables \ref{tab:table6} and \ref{tab:table7}, it can be observed that for large data sets, the combination of AWELE and SMIPR not only improve accuracy but also training efficiency.

\subsection{Efficiency Comparison} \label{ssec:efficiency}
The training time of each approach on the three data sets is shown in Table \ref{tab:table7} and their trends (based on $log_{10}$) are drawn in Fig \ref{fig:times} for easier comparison. Obviously, MIMLfast is the most efficient one. However, as illustrated in Tables \ref{tab:table3} and \ref{tab:table4}, MIMLfast does not work well in the four MIML metrics (HL, OE, RL, AP) because MIMLfast only employs a simple linear classifier and lacks preprocessing the raw images. Although such a framework greatly improves efficiency, it does not work well on the raw images. MIMLBoost is most time-consuming, followed by MIMLmiSVM and MIMLkNN. As shown in Table \ref{tab:table7}, the advantage of our proposed BMIML is obvious. On \textit{ODR}, MIMLBoost can obtain results in 72 hours for the two smallest subsets only, while MIMLmiSVM can handle only 2000 samples. In contrast, BMIML takes only 19 hours even for the largest size (120K). On NIH, MIMLBoost and MIMLmiSVM fail to obtain any result in 72 hours even with the smallest size, while MIMLkNN and MIMLSVM cannot work when the data size reaches 90k, but BMIML can still work well and efficiently. On the largest data (\textit{NIH\_120K}), the advantage of BMIML is even more obvious. Except for MIMLfast, none of the existing methods can deal with large data sets faster than BMIML. In Table \ref{tab:table7}, both MIMLfast and AWELE can achieve high efficiency, but when the data size reaches 30k, the efficiency of AWELE decreases significantly. As observed in Tables \ref{tab:table4}, \ref{tab:table6} and \ref{tab:table7}, MIMLfast is sensitive to the number of instances, and AWLEL is sensitive to the number of bags. In other words, the time cost is not only related to the number of bags but also to the number of instances in each bag.

\section{Conclusion} \label{sec:conclusion}
In this paper, an accurate and efficient BMIML framework was successfully developed, which is suitable for multi-label image classification in medical scenarios. The proposed framework consists of three novel modules i) auto-weighted label enhancement learning (AWELE), ii) scalable multi-instance probabilistic regression (SMIPR), and iii) interactive decision optimization (IDO). AWELE fully takes into account the inter-correlations of the bags, instances, and labels from the training sample, leading to more effective classification. Compared to the existing indirect methods, SMIPR utilizes the inter-instance correlations directly which can reduce the information loss incurred during the conversion process so that it is more effective and efficient than existing indirect methods. IDO works as a bridge to interactively combine and optimize the results from AWELE and SMIPR. Therefore, an interactive end-to-end single network for MIMIL becomes possible, which has never been done in the literature. Extensive experiments were conducted on several real-world medical image databases. The results demonstrate that the proposed BMIML is: i) highly effective (improved by up to $2\%-40.7\%$ on AP) under the four-evaluation metrics (HL, OR, RL, AP) than other state-of-the-art MIML algorithms; ii) significantly more efficient (about $16.56\%-99.99\%$ faster) than most existing algorithms while dealing with large data sets (except for MIMLfast, which is with very poor accuracy). In the future, we will try to employ other kinds of images rather than medical images only.

\bibliographystyle{Reference/IEEEtran}
\bibliography{Reference/IEEEabrv,Reference/IEEEexample}

\newpage
\vfill
\end{document}